\documentclass{article}

\usepackage{arxiv}
\usepackage{graphicx}
\usepackage[utf8]{inputenc} 
\usepackage[T1]{fontenc}    
\usepackage{hyperref}       
\usepackage{url}            
\usepackage{booktabs}       
\usepackage{amsfonts}       
\usepackage{nicefrac}       
\usepackage{microtype}      
\usepackage{lipsum}
\usepackage{color}
\usepackage{outlines}

\newcommand{\remove}[1]{}

\newcommand{\spokennametovec}{\emph{SpokenName2Vec }}

\usepackage{booktabs}
\usepackage{multirow}

\title{How Does That Sound? Multi--Language SpokenName2Vec Algorithm Using Speech Generation and Deep Learning}

\author{
  Aviad Elyashar, Rami Puzis, and Michael Fire \\
  Department of Software and Information Systems Engineering,\\
  Ben-Gurion University of the Negev, Beer-Sheva, Israel \\
  \textit{aviade@post.bgu.ac.il, \{puzis, mickyfi\}@bgu.ac.il} \\
}

\begin{document}
\maketitle

\begin{abstract}

Searching for information about a specific person is an online activity frequently performed by many users.
In most cases, users are aided by queries containing a name in Web search engines for finding their will. 
Typically, Web search engines provide just a few accurate results associated with a name-containing query.
Most existing solutions for suggesting synonyms in online search are based on pattern matching and phonetic encoding, however very often, the performance of such solutions is less than optimal.
In this paper, we propose \emph{SpokenName2Vec}, a novel and generic algorithm which addresses the similar name suggestion problem by utilizing automated speech generation, and deep learning to produce spoken name embeddings.
These sophisticated and innovative embeddings capture the way people pronounce names in any language and accent.
Utilizing a name's pronunciation can be helpful for both differentiating and detecting names that sound alike, but are written differently.
The proposed approach was demonstrated on a large-scale dataset consisting of 250,000 forenames and evaluated using a machine learning classifier and 7,399 names with their verified synonyms.
The performance of the proposed approach was found to be superior to 10 other algorithms evaluated in this study, including well used phonetic encoding and string similarity algorithms, and two recently proposed algorithms (e.g., Name2Vec and GRAFT).
The results obtained suggest that the proposed algorithm could serve as a useful and valuable tool for solving the problem of synonym suggestion.

\end{abstract}


\section{Introduction}
\label{sec:introduction}

In information systems, searching for information about a specific individual is a frequently performed activity~\cite{yang2006web}; for example, retrieving a patient's electronic medical record from a medical records system~\cite{pfeifer1996retrieval} and searching for a research paper online by the author's name or a news article by a journalist's name are daily tasks performed using individuals' names. 
Names are also the focus of the online search, and individuals' reliance on names, as reflected in search engine queries, is steadily increasing. 
For example, in 2004, 30\% of all search engine queries provided by users included personal names~\cite{guha2004disambiguating}.
A decade later, in 2014, one billion names were used in Google search engine queries each day~\cite{GoogleYourself}.

While the use of personal names in online search has increased, the results retrieved from Web search engines has not kept pace~\cite{jansen2000real}. 
Leading online search engines retrieve suboptimal results in response to searches for a person's name~\cite{spink2002sex}. 
These poor results created a new customer need~\cite{organicweb} which has been fulfilled by companies, such as Pipl\footnote{https://pipl.com/}
and ZoomInfo,\footnote{https://www.zoominfo.com/}
which have dedicated their efforts towards providing information about specific people.
Despite these new services, in many cases, users experience difficulty when selecting the exact name to search for or the correct form of a name when formulating a name-containing query. 
Therefore, searching for people by name online remains a challenging problem.

There are several reasons for the poor search engine performance for queries containing names.
First, unlike words, which, in most cases, have a single correct spelling, there are several legitimate variations for a given name~\cite{christen2006comparison}.  
Second, there are cases in which a name changes over time due to the use of a nickname, marriage, religious conversion (e.g., from Lewis Alcindor Jr. to Kareem Abdul Jabbar), or gender reassignment.
Third, many names are heavily influenced by a person's cultural background~\cite{christen2006comparison}. 
For example, the English forename of Anthony has several variations in other languages: Antoine (French), Antonius (Ancient Roman), Anton (Russian), and Antonio (Spanish)~\cite{anthony_behindthename}.
The detection of aliases for people also poses a  challenge;
for instance, the nickname of Kobe Bryant, the famous basketball player, is the ``Black Mamba.''
Therefore, finding a match for a name is more difficult than it is for general text~\cite{borgman1992getty}.

Today, techniques used for name matching and the retrieval of similar names are mainly based on pattern matching and phonetic encoding~\cite{christen2006comparison}.
For example, in the context of names,  phonetic encoding algorithms (e.g., Soundex) encode a given name into plain-text code that reflects the way people pronounce the name.
This plain-text code assists in finding similar names in cases in which the code for two different names is identical (e.g., Smith and Smyt).
However, the performance of these algorithms has been poor~\cite{friedman1992tolerating}.

In recent decades, there has been a data science revolution resulting in the development of products and services that utilize machine and deep learning algorithms to help people in various aspects of modern life, for example, searching for information on the Internet, filtering spam email, image recognition, etc.~\cite{grapov2018rise}
These advanced algorithms, which are capable of learning from a large set of examples, were found much more effective and robust than those that were designed using explicitly specifying rules~\cite{gulshan2016development}. 
For example, \emph{Word2Vec}~\cite{word2vec_paper} is a deep learning-based model that utilizes large-scale text to transform words into continuous vector space representations (also known as word embeddings). 
These fixed-dimensional vector representations were found to have semantic meaning, which can be used for many natural learning processing (NLP) tasks, such as text classification, word similarity, and more.

Inspired by \emph{Word2Vec}, we propose a novel and generic approach that leverages the power of human speech and deep learning to address several issues associated with names, such as synonym suggestion and record linkage.
The proposed \spokennametovec approach is an innovative multi-language algorithm that uses names, languages, accents, and automated speech generation to produce spoken name embeddings.
These novel embeddings capture linguistic and acoustic content which is used to detect names that sound alike.
In contrast to phonetic encoding algorithms, such as Soundex and Double Metaphone, which represent names with plain-text code, the proposed algorithm utilizes neural networks to create more advanced name representations, viewed as fixed-length space vectors. 
The continuous vector space representation for names is based on the way humans pronounce names in any language, with any accent (e.g., American and British English).
To the best of our knowledge, we are the first to represent names using spoken name embeddings.

In this paper, we demonstrate the proposed algorithm on the task of suggesting synonyms associated with a given name, a common task required of search engines today.
The \spokennametovec  algorithm consists of five phases: 
(1) the name collection phase, in which we collect names; 
(2) the speech segment generation phase, in which we generate spoken names based on the given name, \textit{targeted language}, and \textit{accent};
(3) the feature extraction phase, where we extract audio features which serves as a continuous vector space representation for each name; 
(4) the classification phase, in which a machine learning classifier is used to classify candidates that sound like the given name;  
and (5) the last phase, in which candidates are filtered according to a predefined threshold (the remaining candidates serve as synonyms for the given name).

In our evaluation, the performance of the proposed algorithm is compared with the performance of other the state-of-the-art machine and deep learning algorithms. 
The performance was evaluated using the Behind the Name dataset with over 7,300 forenames and over 37,000 synonyms.
We show that the proposed \spokennametovec algorithm outperforms all other algorithms evaluated, including commonly used phonetic encoding and string similarity algorithms, as well as novel algorithms suggested more recently (e.g., the GRAFT~\cite{elyashar2019runs} and Name2Vec~\cite{foxcroft2019name2vec} algorithms) in terms of the average accuracy, F1, and precision@5 and precision@10 measures.
For example, \spokennametovec configuration which utilized spoken names in the Italian accent obtained an average accuracy score of 0.151 in contrast to the GRAFT and Double Metaphone algorithms which obtained average scores of 0.114 and 0.068, respectively.

The remainder of this paper is organized as follows:
In Section~\ref{sec:related_work}, we provide a brief overview of related work focused on issues similar to those addressed in this study.
Section~\ref{sec:methods} presents the \spokennametovec framework. 
We provide a detailed description of the datasets used in this study in Section~\ref{sec:data}.
In Section~\ref{sec:experimental_setup}, we review the experimental setup, and in Section~\ref{sec:results}, we present the performance (for the task of suggesting synonyms) of the proposed algorithm and the other algorithms evaluated.
In Section~\ref{sec:discussion}, we discuss the results obtained, and our conclusions and future directions are provided in Section~\ref{sec:conclusion}.

\section{Background}
\label{sec:related_work}

In the subsections that follow, we provide the necessary background on this study and review related work. 
In Section~\ref{sec:speech_and_name_representation}, we provide brief background information related to speech, including an overview of the existing automated mechanism for generating speech automatically.
In Section~\ref{sec:speech_and_name_representation}, we present existing representations for speech and names.
Our proposed spoken name embedding relies on the extraction of audio features from audio segments, and Section~\ref{sec:audio_feature_extraction} presents the mechanism used to extract these features. 
Then, in Section~\ref{sec:string_similarity_metrics}, we provide a brief overview of a few well-known string similarity algorithms and phonetic algorithms (see Section~\ref{sec:phonetic_encoding_algorithms}) which our proposed algorithm is compared to when evaluating its performance.
Lastly, in Section~\ref{sec:related_name_suggestion}, we review previous studies that focused on suggesting similar names associated with a given name.


\subsection{Speech and Name Representation}
\label{sec:speech_and_name_representation}

In this paper, we propose a novel representation for names, which uses automated speech and deep learning to deal with problems associated with names, such as similar name suggestion~\cite{elyashar2019runs} and record linkage~\cite{foxcroft2019name2vec}.
Most of the well-known approaches aimed at handling these problems emphasize character or word similarities (e.g., the edit distance string similarity algorithm).
In contrast to these approaches, \spokennametovec addresses these problems by utilizing the power of speech to find similar names.
In addition to the use of speech for conveying ideas and expressing feelings~\cite{tiwari2012voice}, it has been helpful for other related tasks, such as voice recognition~\cite{klevans1997voice}, speaker recognition~\cite{jayamaha2008voizlock}, analyzing human behavior~\cite{lepine1998predicting}, Internet communication~\cite{goode2002voice}, name suggestion~\cite{hall1980approximate}, and more. 
Often there are several variations of names (e.g., Smith, and Smyt), which are written differently but pronounced the same. 
Focusing on the way names are pronounced instead of how they are written can be a salient advantage for the detection of synonyms.    
For this, we use open-source and publicly available services for generating automated speech, e.g., the Text2Speech website\footnote{https://www.text2speech.org/} and Google Text-to-Speech.\footnote{https://cloud.google.com/text-to-speech} 

The data science revolution of the last decade has resulted in the development of many products that use machine and deep learning algorithms, including products for filtering spam images, image recognition, and more. 
Among the pioneers of these algorithms were Mikolov et al.~\cite{word2vec_paper}, who in 2013
introduced Word2Vec's architecture for word embedding.
Word2Vec is a generic method encompassing two representation learning models: continuous bag of words (CBOW) and skip-gram.
Both models are simple feed-forward neural network architectures that are used for computing continuous vector representations of words from very large datasets.
The vector representations of words learned by Word2Vec held promise for maintaining semantic meanings, a trait that is useful for various natural language processing (NLP) tasks, such as text classification~\cite{lilleberg2015support}, information retrieval~\cite{ganguly2015word}, etc.
In 2014, Le and Mikolov~\cite{le2014distributed} extended the Word2Vec methodology and suggested Doc2Vec, a fixed-dimensional vector representation for sentences and documents using a paragraph vector.
This additional vector remembers the context or the topic of each paragraph, which was shown to be useful for capturing the semantics of paragraphs, sentences, and documents.
In recent years, many researchers, inspired by the novel Word2Vec algorithm, have suggested utilizing the power of representation learning on various domains that are not necessarily related to NLP.
Examples of the models proposed include Node2Vec~\cite{grover2016node2vec}, App2Vec~\cite{ma2016app2vec}, Song2Vec~\cite{rosssong2vec}, and Emoji2Vec~\cite{eisner2016emoji2vec}, and more.

In 2018, Chung and Glass~\cite{chung2018speech2vec} proposed Speech2Vec, a speech version of Word2Vec.
To train their model, they used LibriSpeech, a corpus of 500 hours of English speech, to learn Speech2Vec embeddings.
They compared their model with the classic Word2Vec algorithm on word similarity tasks. 
Later that year, Chung et al.~\cite{weng2018towards} tested the Speech2Vec models on the task of speech-to-text translation.
In 2019, Haque et al.~\cite{haque2019audio} proposed spoken sentence embeddings.
Their results demonstrated that the proposed spoken sentence embeddings outperformed phoneme and word-level baselines on speech and emotion recognition tasks. 
In the same year, Foxcroft et al.~\cite{foxcroft2019name2vec} presented \emph{Name2Vec}, a method for name embeddings that employs the Doc2Vec methodology, where each surname is viewed as a document, and each letter constructing the name is considered a word.
They demonstrated the task of record linkage by training a few name embedding models on a dataset containing 250,000 surnames and tested their model on 25,000 verified name pairs from Ancestry.com. 
They used the Ancestry Records dataset as positive samples and other 25,000 random name pairs as negative samples.
The authors concluded that the name embeddings generated can predict whether a pair of names match.

\subsection{Audio Feature Extraction}
\label{sec:audio_feature_extraction}

In the feature extraction phase, in order to analyze the audio data obtained by generating spoken names and produce spoken name embeddings, we extract audio features using open-source frameworks that specialize in extracting features from audio files.
Such frameworks are mainly used for tasks like audio event recognition and surveillance, speech recognition, and music information retrieval~\cite{giannakopoulos2015pyaudioanalysis}; 
examples of libraries and frameworks for this include Yaafe,\footnote{http://yaafe.sourceforge.net/} librosa,\footnote{https://github.com/librosa/librosa} PyCASP,\footnote{https://github.com/egonina/pycasp} Bob,\footnote{http://idiap.github.io/bob/} pyAudioAnalysis~\cite{giannakopoulos2015pyaudioanalysis}, and Turi Create's sound classifier~\cite{sound_classifier}. 

In this study, we extract audio features using two frameworks: the Turi Create sound classifier and pyAudioAnalysis.
With Turi Create~\cite{sound_classifier}, this phase includes the following signal processing steps to transform the audio segments into convenient data for use as neural network input:
First, the raw audio frequency signals are transmitted into a series of digital numbers (from 1 to -1) using pulse code modulation (PCM)~\cite{shorter1972application}.
All of the signals are re-sampled to 16,000 samples per second.
The data is then divided into several overlapping windows.
For each window, the Hamming window, a mathematical function that is zero-valued outside of some chosen interval, is applied;
this function window is widely used in digital signal processing applications~\cite{podder2014comparative}.
The power spectrum is calculated using fast Fourier transformation, and finally Mel Frequency filter banks are applied and the natural logarithms of all of the values are used as features.

The pyAudioAnalysis framework was implemented by Giannakopoulos~\cite{giannakopoulos2015pyaudioanalysis} in 2015.
This framework includes the calculation of 11 types of audio features, including zero crossing rate; energy;  entropy of energy; spectral centroid; spread, entropy, flux, and rolloff; Mel-frequency cepstral coefficients (MFCCs), chroma vector, and deviation.


\subsection{String Similarity Algorithms}
\label{sec:string_similarity_metrics}

To evaluate \emph{SpokenName2Vec}, we compared its results with the results of string similarity algorithms.
These well-known algorithms 
have usually been used to match individuals or families of samples for tasks, such as measuring the coverage of a decennial census or combining two databases, such as tax information and population surveys~\cite{cohen2003comparison,casanova2007database}.
Such algorithms determine the similarity of two given strings by measuring the ``distance'' between the strings.
Two strings that are found similar by these algorithms are considered related. 
In this study, we evaluate the performance of the following string similarity functions:

\textbf{Damerau-Levenshtein Distance.} 
The \emph{Damerau-Levenshtein distance} was developed in 1964 by Damerau~\cite{damerau1964technique}.
To transform a given word to another, this string algorithm measures the minimal number of four different types of editing operations: insertion, deletion, permutation, and replacement.

\textbf{Edit Distance.}
The \emph{edit distance}, also known as the \emph{Levenshtein distance}, was developed two years later by Levenshtein~\cite{levenshtein1966binary}.
This similarity string algorithm measures the minimal number of operations required to transform one word into an other~\cite{levenshtein1966binary}. 
These operations are insertions, deletions, and substitutions of a single character.
For example, the \emph{edit distance} between the names John and Johan is one.

\textbf{Jaro-Winkler Distance.}
This string distance was developed in 1995 by Jaro and Winkler~\cite{jaro1995probabilistic,jaro1989advances,winkler1999state}.
This metric was intended primarily for short strings like personal surnames~\cite{cohen2003comparison}.
It is based on the number and order of the common characters between two given strings~\cite{cohen2003comparison}.
The lower the Jaro–Winkler distance for two strings is, the more similar the strings are.
This is normalized such that zero means an exact match and one means there is no similarity.
In this study, we used the Jaro–Winkler similarity metric, which is the inversion of the distance described above.

\subsection{Phonetic Encoding Algorithms}
\label{sec:phonetic_encoding_algorithms}

Other algorithm families whose performance we compare to \emph{SpokenName2Vec}'s performance are the phonetic encoding algorithms.
These algorithms are methods that transform a given word into a code according to the way the word is pronounced. 
These algorithms are commonly used for spelling suggestion~\cite{uzzaman2004bangla}, entity matching~\cite{cohen2003comparison,peled2013entity}, and searching for names in websites~\cite{khan2017application} or databases~\cite{patman2001soundex}.
In this paper, we evaluate the following phonetic encoding techniques: Soundex, Metaphone, Double Metaphone, the New York State Identification and Intelligence System Phonetic Code (NYSIIS), and the match rating approach (MRA).

\textbf{Soundex.} Devised over a century ago by Russel and O’Dell, the Soundex algorithm is one of the first phonetic encoding techniques~\cite{hall1980approximate}.
Given a name, it provides a code that reflects how it sounds when spoken.
It keeps the first letter in a given name and reduces all of the remaining letters into a code of one letter and three numbers.
Vowels and the letters \textit{h} and \textit{y} are converted to zero. 
The letters \textit{b}, \textit{f}, \textit{p}, and \textit{v} are converted to one.
The letters \textit{c}, \textit{g}, \textit{j}, \textit{k}, \textit{q}, \textit{s}, \textit{x}, and \textit{z} are converted to two. 
The letters \textit{d} and \textit{t} are converted to three, while \textit{m} and \textit{n} are converted to five.
The letter \textit{l} is converted to four, and \textit{r} is converted to six.
The final code includes the original first letter and three numbers.
Codes that are generated for longer names are cut off, whereas shorter codes are extended with zeros.
For example, the Soundex code for the name Robert is R163.  

\textbf{Metaphone.} The Metaphone algorithm was developed in 1990 by Lawrence Philips~\cite{philips1990hanging}. 
It is an improvement over Soundex, because the words are encoded to a representation so that they can be combined into a group despite minor differences~\cite{binstock1995practical}.
This algorithm assumes English phonetics and works equally well for forenames and surnames~\cite{pimpalkhute2014phonetic}.
It widely used in spell checkers, search interfaces, genealogy websites, etc~\cite{khan2017application}.
The Metaphone code for the forename Robert is RBRT.  

\textbf{Double Metaphone.} The Double Metaphone algorithm was developed almost two decades ago by Lawrence Philips~\cite{philips2000double}.
A variation of the Metaphone algorithm, the Double Metaphone, retrieves a code that consists solely of letters.
As opposed to the previous two algorithms, the Double Metaphone also attempts to encode non-English words (European and Asian names).
Moreover, unlike all other phonetic algorithms, it returns two phonetic codes.
For example, the Double Metaphone code for the forename Jean is JN and AN. 

\textbf{NYSIIS.} This phonetic encoding algorithm also returns a code that consists solely of alphabetic letters~\cite{borgman1992getty}, however it preserves the vowels' positions in a given name by converting all of the vowels to the letter `A'~\cite{de1986guth}. 
For example, the NYSIIS code for the forename Robert is RABAD.

\textbf{Match Rating Approach (MRA).}
This phonetic encoding algorithm was developed by Gwendolyn Moore in 1977~\cite{moore1977accessing}. 
The algorithm includes a small set of encoding rules, as well as a more lengthy set of comparison rules.
For example, the returned code for the forename Robert is RBRT.

\subsection{Similar Name Suggestion Algorithms}
\label{sec:related_name_suggestion}

In the late two decades, several studies have confronted the problem of similar name suggestion.
In 1996, Pfeifer et al.~\cite{pfeifer1996retrieval} compared the differences in the performance of a few known phonetic similarity measures and exact match metrics for the task of improving the retrieval of names.
For the evaluation process, the authors collected surnames manually from a few sources, such as the TREC collection~\cite{harman1992overview}, the CACM collection from the SMART system~\cite{buckley1985implementation}, the phonebook of the University of Dortmund, Germany, and author names from a local bibliographic database.
They combined all of the surnames into the COMPLETE dataset, which includes approximately 14,000 names.
They determined the queries for this dataset as follows: 
First, they chose 90 names randomly from the COMPLETE dataset.
Second, for each of the 90 queries, they manually determined the relevant names.
They reported that an information system based on phonetic similarity measures, such as Soundex, and variations of phonetic algorithms outperform exact match search metrics in the task of searching for synonyms.

In 2010, Bollegala et al.~\cite{bollegala2010automatic} suggested a method for extracting aliases for a given personal name based on the Web; for example, the alias of the term ``fresh prince'' is Will Smith. 
They proposed a lexical pattern-based approach for extracting aliases of a given name using snippets returned by a Web search engine. 
Then, they defined numerous ranking scores to evaluate candidate aliases using three approaches: lexical pattern frequency, word co-occurrences in an anchor text graph, and page counts on the Web.
Their method outperformed numerous baselines, achieving a mean reciprocal rank of 0.67.
There are a few differences between this study and ours.
First, our study focuses on the task of suggesting similar names that sound like a given name, while 
Bollegala et al. focused on suggesting aliases. 
An alias can be very different from a given name.
For example, the aliases of the famous basketball players, LeBron James and Earvin Johnson, are ``the King'' and ``Magic,'' respectively. 

Recently, Elyashar et al.~\cite{elyashar2019runs} proposed GRAFT, a novel approach for suggesting synonyms using the construction and analysis of digitized family trees.
Using a large-scale online genealogical WikiTree dataset, the authors constructed a graph based on names derived from digitized family trees. 
Utilizing this very large graph, they suggested synonyms by searching for the given name in the graph and traversed from it to collect the suggested candidates.
Next, they applied four ordering functions determining the order of the suggested names.
GRAFT outperformed phonetic and string similarity algorithms for the task of suggesting synonyms.
In contrast to this approach, which utilizes historical knowledge to detect synonyms based on ancestors, the main advantage of \spokennametovec is its ability to detect many synonyms that sound like the given name, without the need for historical data which may or may not be available. 

In addition to studies aimed at developing techniques for suggesting synonyms, several companies have emerged to address the task of using names to find people online in response to the growing need of Internet users to find people online and the poor results provided by the largest search engines~\cite{organicweb}. 
Among them are Pipl, which utilizes names to search for the real person behind online identities~\cite{pipl_about_us}, and
ZoomInfo~\cite{zoominfo_about_us}, which provides company or organizational oriented information for a searched name.
According to ZoomInfo, 
their database includes 67 million emails and 20 million company profiles.

Other free online services include: PeekYou,\footnote{https://www.peekyou.com/} 
a people search website that collects and combines content from online social networks, news sources, and blogs to help retrieve the online identity of American users and TruePeopleSearch,\footnote{https://www.truepeoplesearch.com/} which helps find people by name, phone number, or address.
Websites, such as TruthFinder\footnote{https://www.truthfinder.com/} and BeenVerified\footnote{https://www.beenverified.com/} provide background checking services for people. 
These services can help reconnect Americans with their friends and relatives, as well as provide a way to look up criminal records online.


\section{Methods}
\label{sec:methods}

In this paper, we present \emph{SpokenName2Vec}, a novel and generic deep learning algorithm utilizing multi-language automated speech 
for various tasks related to names. 
In this section, we present the steps of the proposed algorithm and demonstrate its effectiveness for the task of suggesting names that are similar to a given name.
Similarly to the phonetic encoding algorithms (e.g., Soundex and Double Metaphone), the proposed \spokennametovec algorithm transforms a given name into a single representation.
However, in contrast to those methods, after encoding text into a simple plain-text code, the proposed algorithm generates a fixed-dimensional vector representation derived from an audio segment expressing the way people pronounce a given name in a given language and accent. 
This results in a deep neural network-based model, which takes into account the given name, as well as the language and accent.
This model is much more sophisticated than other algorithms evaluated for the task of suggesting synonyms, and its ability to detect names that sound alike but are written differently is notable.

\subsection{Multi-Language SpokenName2Vec}
\label{sec:name2vec}

The proposed algorithm consists of the following five steps (see Figure~\ref{fig:proposed_method}):

\begin{figure*}[h!]
\centering
\includegraphics[scale=0.6]{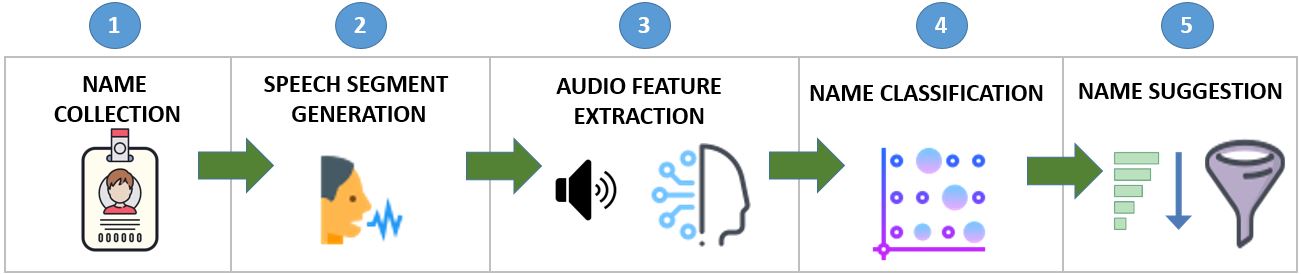}
\caption{Overview of the \emph{SpokenName2Vec}'s steps.}
\label{fig:proposed_method}
\end{figure*}

\begin{enumerate}
\item \textbf{Name Collection.} 
To apply the proposed innovative \emph{SpokenName2Vec} algorithm, a dataset of names is required.
Names can be obtained from genealogical websites, online social networks, other designated websites, and other services. 
Of course, a preprocessing step is required to remove noisy and unnecessary data from these names, such as short abbreviations, honorific titles, etc.

\item \textbf{Speech Segment Generation.} 
After obtaining a collection of names, audio segments are generated, reflecting how humans say each name according to a given language and accent.
The generation of audio segments is performed using tools that transform text for a given name into speech segments automatically. 
This step is generic, i.e., we can transfer text to speech by selecting any of the languages used by tool, along with the associated accent, to generate the speech segment.
The speech segment generation step results in a collection of speech segments reflecting the names collected in Step 1 and spoken according to a target language and accent.

\item \textbf{Speech Segment-Based Feature Extraction.} 
In this step, each speech segment generated is transformed into a  fixed-dimensional vector space representation using deep learning implemented by an artificial neural network-based model.
This sophisticated representation, which consists of several dimensions, obtains linguistic and acoustic content concerning the spoken name. 
For this, we use state-of-the-art algorithms to transform an audio segment into a fixed-dimensional vector representation.
The resulting vectors, also known as spoken name embeddings serve as features for each of the given names; these features are used in the next steps. 

\item \textbf{Name Classification.}
Utilizing the extracted speech segment-based features, we use supervised machine learning classifiers to suggest synonyms associated with a given name. 
This step is generic and compatible with many classifiers, such as classification-based nearest neighbor classifiers, classifiers that apply kernel functions, and others. 
For a given name, this step results in at most $k$ candidate names suggested (based on the classifier's predictions) as synonyms, as well as a confidence score associated with each candidate reflecting the candidate's likelihood of being found as a correct synonym for the given name.

\item \textbf{Name Suggestion.}
The name suggestion step consists of two actions: filtering candidates and applying an order function. 
First, the candidates that sound different than the given name are filtered, by determining a threshold.
The filtering action is performed using a confidence score provided by the chosen classifier.
Therefore, we order all of the candidates associated with the given name according to the confidence score provided, where the candidates with the highest confidence score are placed first.
A threshold is then determined;
as a rule of thumb, the threshold should be set such that all of the candidates the classifier is not certain about are removed. 
Second, we order the remaining candidates using an ordering function. 
In this step, various ordering functions can be used, including Damerau-Levenshtein, edit distance, and more. 

\end{enumerate}

\section{Data Description}
\label{sec:data}

To evaluate the proposed algorithm, we used three datasets: the WikiTree, Spoken Name, and Behind the Name datasets.
The WikiTree dataset includes names from previous generations.
The Spoken Name dataset is a collection of audio segments taken from an automated speech generation process and includes a name, as well as its language and accent.
The Behind the Name dataset provides the ground truth for evaluating the performance of the \spokennametovec algorithm and the other algorithms evaluated.

\subsection{WikiTree Dataset}
\label{sec:wikitree_dataset}

We used genealogical records available on the WikiTree website~\cite{Wikitree_dump}.
WikiTree is an online genealogical website founded in 2008 by Chris Whitten~\cite{wikitree}.
Its main aims are to provide a framework and genealogical sources for creating an accurate single family tree and make genealogy free and accessible worldwide.
As of February 2020, WikiTree had over 680,000 registered users and maintained over 22 million profiles~\cite{wikitree}.
Many of these profiles contain specific details about each individual, such as their full name, nickname, gender, birth and death dates, children's profiles, etc.
The massive WikiTree dump we worked with includes more than 17 million profiles and over 250,000 unique forenames.

\subsection{Spoken Name Dataset}
\label{sec:spoken_name_dataset}

This dataset is a collection of audio segments (WAV files) of names pronounced by an automated text-to-speech framework. 
We used the Google Text-to-Speech Python library (gTTs)~\cite{gtts} which supports multiple languages and accents.
For each name in the WikiTree dataset, we generated a speech segment reflecting how people pronounce it in a target language and accent.
The Spoken Name dataset consists of six different languages: American English, French, Spanish, Chinese, Russian, and Italian.  
Each language includes 250,038 WAV files associated with the names in the dataset.


\subsection{Behind the Name Dataset}
\label{sec:behind_the_name_dataset}

To evaluate the performance of the proposed algorithm and compare it to other methods, we needed a ground truth dataset.
Therefore, we generated the following ground truth dataset by combining the information included in the WikiTree dataset with the data on the Behind the Name website~\cite{behindthename}. 
This website was founded in 1996 by Mike Campbell in order to study various aspects of names~\cite{behindthename_info}. 
It contains names from all cultures and time periods, as well as mythological and fictional names.
Currently, the website contains 22,263 names.

The creation of the ground truth dataset was performed, as follows:
First, we extracted all the distinct forenames in the WikiTree dataset with a length greater than two letters (to avoid honorific titles).  
From the over 17 million profiles available at the time of this research, we extracted 250,038 unique forenames.   
Using the public service application programming interface (API) provided by Behind the Name, we collected synonyms for the unique forenames in the WikiTree dataset.
For example, for the given name of Ed, we collected Eddie, Edgar, Edward, Ned, Teddy, etc.~\cite{ed_behindthename}.
For the given name of Elisabeth, we retrieved Eli, Elisa, Ella, Elsa, Lisa, and Liz~\cite{elisabeth_behindthename}.
In total, 37,916 synonyms were retrieved for the 7,399 distinct names in the WikiTree dataset.
The names that provided the greatest number of synonyms were Ina, Nina, and Jan with 127, 119, and 92 synonyms, respectively. 
On average, the Behind the Name dataset contains 5.12 synonyms for a given forename.


\section{Experimental Setup}
\label{sec:experimental_setup}

\subsection{Setting Experimental Parameters}

In this study, we conducted experiments aimed at answering two research questions: 
(1) \textit{Is the proposed \spokennametovec vector representation valid and useful?} 
(2) \textit{Can the proposed algorithm's performance be improved by utilizing specific languages and accents in the speech segment generation step?} 

\subsubsection{Vector Representation Validation}
To answer the first research question regarding performance, we evaluated the performance of the proposed \spokennametovec algorithm for the task of suggesting similar names for a given name by conducting a large-scale experiment as follows: 
First, we obtained a collection of forenames. 
For this, we used the WikiTree dataset (see Section~\ref{sec:wikitree_dataset}). 
As mentioned earlier, preprocessing was required; therefore, we cleaned the forenames by removing short names that contained less than three characters (see Section~\ref{sec:name2vec}, Step 1). 

Second, we used the gTTS library~\cite{gtts} 
to transform the forenames collected into speech segments reflecting the names as expressed by humans in their native tongue according to four different languages and accents: American English, French, Spanish, and Italian.  
In total, for each language, 250,038 WAV files were generated. 

Then, for each speech segment representing a name, we extracted audio features using two open-source frameworks: Turi Create's sound classifier~\cite{sound_classifier} and pyAudioAnalysis~\cite{giannakopoulos2015pyaudioanalysis}.
In total, for each name, we generated 12,288 audio features using Turi Create and 136 audio features using pyAudioAnalysis, which served as a fixed-dimensional vector representation for each name.

Next, using the audio features obtained, together with the K-nearest neighbors (KNN) classifier, we selected the $k$ nearest neighbors as candidates to be suggested as synonyms for each given name, where $k = 10$. 

For each given name, we first sorted the $k$ candidate names based on their Euclidean distance from the vector representation of the given name and removed forenames for which the vector representation's distance was greater than one (for the  audio features extracted using Turi Create's sound classifier) and greater than zero (for the audio features extracted using pyAudioAnalysis).\footnote{The reason for the differences in the thresholds is related to the audio feature extraction step. 
The Euclidean distance between the given name and its candidates for the 12,288 audio features extracted using Turi Create was diverse (from zero to 12), whereas when utilizing the 136 audio features extracted using pyAudioAnalysis, all of the Euclidean distance scores ranged from zero and one.} 
Then, we used the edit distance as an ordering function (i.e., the edit distance score is calculated between each given name and the remaining candidates). 
Finally, we sorted the candidates in ascending order according to the edit distance score, and these served as the suggested synonyms for each given name.  
To measure the validity of the proposed representation, we evaluated it for the task of suggesting synonyms using objective performance metrics, such as accuracy, F1, precision, and recall.




\subsubsection{Language and Accent Comparison}

To answer the second research question, we conducted an empirical experiment in which we evaluated the performance of \spokennametovec on a specific language and associated accent for the task of suggesting names that are commonly used in countries and regions that mainly speak the language. 
In other words, in order to improve the performance, we analyzed whether a selected language and accent should be taken into account.
To do this, we conducted the following experiment:
First, we used the Behind the Name dataset. 
For each name, we utilized the Behind the Name website to identify where (countries and regions) each given name is commonly used; for example, according to the website, the forename of Alfredo is commonly used in Italy, Spain, and Portugal\footnote{https://www.behindthename.com/name/alfredo}).


Second, we applied five configurations of the \spokennametovec algorithm using Turi Create's sound classifier and five different languages: English, French, Spanish, Italian, and Russian.

Then, for each configuration reflecting a language and accent, we selected the names that exist in the ground truth which are also commonly used in specific countries and regions; for example, we defined English names as names whose usage, according to Behind the Name website was English (we also considered sub-categories used in the website, such as general, modern, rare, and archaic). 
We also included Australian, British, New Zealand, and American names as English, as well as Hispanic, African American, and Anglo-Saxon names.
The same process was performed for French, Italian, Russian names.
For Spanish names, we also included names that are commonly used in Latin America.



Next, for each given name, we searched for the top 10 most similar names according to each language and suggested them as synonyms.
Finally, to evaluate the performance of the proposed configuration, we identified the correct synonyms among the suggested names. 
To evaluate performance, we used the precision measure due to its ability to focus on results that are retrieved at the top in the same manner as in Web.

\subsection{Evaluation Process}
\label{sec:evaulation}

To analyze and evaluate the performance of the proposed \spokennametovec algorithm on the task of name suggestion, we evaluated its performance (see Section~\ref{sec:spokenname2vec_evalaution}), as well as the performance of other algorithms used for suggesting synonyms, such as phonetic encoding algorithms (see Section~\ref{sec:phonetic_encoding_algorithms_evaluation}), string similarity algorithms (see Section~\ref{sec:string_similarity_algorithms_evaluation}), and other recently proposed algorithms, such as GRAFT (see Section~\ref{sec:name_based_graph_evaluation}) and Name2Vec (see Section~\ref{sec:name2vec_evaluation}).     
The performance of each of the algorithms was evaluated using the performance metrics of accuracy, F1, precision, and recall. 
For the precision measure, we used the top suggestions provided by each algorithm and calculated the metric of $average\mbox{-}precision@k,$ for $k = 1, 2, 3, 5, 10$.
Similar to the evaluation of search engine ranking, we chose to evaluate the top $k$ suggestions (based on the assumption that our case is similar to the search engine ranking domain, where in most cases, people are only interested in the first page of the results and don't bother to move on to subsequent pages~\cite{brin1998anatomy}).


\subsubsection{Evaluation of \spokennametovec}
\label{sec:spokenname2vec_evalaution}



For each name in the ground truth, we searched for its 10 nearest neighbors using the KNN algorithm, where $k = 10$. 
Then, we filtered some candidate names based on the predefined threshold representing the maximal Euclidean distance between the candidate and given names.
In cases in which the Euclidean distance of the candidate name from the given name is above the threshold, we filtered the candidates.  
Finally, the remaining candidates were placed in ascending order according to their edit distance from the given name (the lower the distance, the higher the similarity). 
The performance was evaluated based on the top $K$ suggestions provided. 


\subsubsection{Comparison to Phonetic Encoding Algorithms}
\label{sec:phonetic_encoding_algorithms_evaluation}

We evaluated the performance of five well-known phonetic algorithms: Soundex, Metaphone, Double Metaphone, NYSIIS, and Matching Rating Approach (MRA) for the task of suggesting synonyms.
The following evaluation process was performed: 
For each given name in the ground truth Behind the Name dataset, we calculated the phonetic code according to the given phonetic encoding algorithm. 
Take, for example, the name of Abraham and the Soundex phonetic algorithm.
First, the name, Abraham, was encoded by Soundex as A165. 
Then, we derived the Soundex phonetic code for all of the other names in the WikiTree dataset.
After that the forenames that have the same phonetic code as Abraham were chosen as candidates; we sorted the candidates according to their edit distance from the given name and retrieved the top $K$ as synonyms.

Unlike phonetic algorithms which produce a single sound code for a given name, Double Metaphone produces two phonetic codes (primary and secondary). 
Therefore, for this algorithm, we collected all of the names that shared the same phonetic code (as either the primary or secondary code) and ordered them according to their edit distance from the given name.

\subsubsection{Comparison to String Similarity Algorithms}
\label{sec:string_similarity_algorithms_evaluation}

We evaluated the performance of two well-known string similarity algorithms (edit distance and Damerau Levenshtein distance).
For this, we measured the given string similarity between each name in the Behind the Name dataset (ground truth) and the candidate name in the WikiTree dataset.
Take, for example, the name of Abraham and the edit distance string similarity algorithm:
First, we calculated the edit distance between each name in the WikiTree dataset and the name of Abraham.
As candidates, we chose just the forenames whose distance from the given name is between one and three.
We limited the edit distance range to be less than or equal to three, since we observed that a larger edit distance value resulted in names that are extremely different than the given name.
In the final step, we sorted the candidates according to their distance. 

With respect to the Jaro-Winkler distance, we set this distance's range between zero to one.
Therefore, we sorted the candidates for a given name in descending order and chose just the top $K$ as synonyms.
Finally, to improve the performance, we sorted the $K$ synonyms according to their edit distance from the given name and retrieved them as synonyms. 

\subsubsection{Comparison to GRAFT}
\label{sec:name_based_graph_evaluation}

To evaluate GRAFT, we followed the steps presented by Elyashar et al.~\cite{elyashar2019runs}, including the construction of the digitized family trees and the graph based on names, using the WikiTree dataset.
For a fair comparison, we applied the hybrid approach (a combination of GRAFT and Double Metaphone) for its ability to suggest synonyms for all names (as same as the proposed \spokennametovec algorithm). 
Like Elyashar et al., we constructed two graphs of names: the first graph for forenames, whereas the latter for surnames.   
For providing the highest performance of GRAFT, the forenames for constructing the first graph were taken from grandparents--grandchildren connections, whereas the surnames for constructing the second graph were taken from parents--children connections. 
The threshold for removing candidates ranged from one to three. 
For both, we applied $NetEDofDMphoneED$, the ordering function that found to maximize the performance of GRAFT.

\subsubsection{Comparison to Name2Vec}
\label{sec:name2vec_evaluation}

We performed two experiments to compare the \spokennametovec and Name2Vec~\cite{foxcroft2019name2vec} algorithms. 
In the first experiment, we evaluated the performance of the Name2Vec approach on forenames using the WikiTree and Behind the Name datasets. 
In the second experiment, we utilized the Ancestry Surnames dataset provided by Foxcroft et al.~\cite{foxcroft2019name2vec} to evaluate \spokennametovec on surnames.

\textbf{Evaluation of Name2Vec on Forenames.} 
First, we used the WikiTree dataset as a data source and trained a Doc2Vec model based on these forenames.
Foxcroft et al. reported that their best model was trained on the Ancestry dataset consisted of 250,000 surnames.
Since the datasets of Ancestry and WikiTree are nearly equal in size (250,000 records), and there are generally not great differences between forenames and surnames, we set the parameters so they were the same as those reported by Foxcroft et al. (640 epochs, 30 dimensions, and a window size of two). 
Next, using the trained model, we collected the 10 most similar candidate names for each forename in the Behind the Name dataset. 
To improve performance, we calculated the edit distance between each given name and its candidates. 
Then, we sorted the candidates in ascending order based on their edit distance score and filtered those candidates with an edit distance score greater than one.\footnote{We tested several predefined thresholds and present the threshold providing the best results in this paper.} 
Finally, we used the remaining candidates as suggested synonyms for the given names that are part of the Behind the Name ground truth dataset.

\textbf{Evaluation of Name2Vec on Surnames.}
In this experiment, we performed the same steps described in the previous paragraph, with two changes:
This time, we trained a Doc2Vec model with the parameters described above on the Ancestry Surnames dataset, which includes 250,000 surnames.
In this experiment, the predefined threshold for surnames was set at those candidates with an edit distance greater than three (instead of one, as was done in the previous experiment).
In this case, we evaluated a few thresholds to maximize the performance of the algorithm and presented the threshold that provided the highest performance (an edit distance ranges from one to three).
Finally, the remaining candidates were evaluated using the Ancestry Records ground truth dataset.

\textbf{Evaluation of SpokenName2Vec on Surnames.}
We evaluated \emph{SpokenName2Vec}’s performance on surnames as follows: 
For each surname in the Ancestry Surnames dataset, we generated a speech segment using gTTS.
Next, we extracted audio features using Turi Create's sound classifier. 
Afterward, using the audio features obtained, together with the KNN classifier, $k$ nearest neighbors were selected as candidates to be suggested as synonyms for each given name, where $k = 10$. 
For each surname, we first sorted the $k$ candidate names according to their Euclidean distance from the generated vector representation of the given name and removed those for which the vector representation's Euclidean distance was greater than one (for the audio features extracted using Turi Create's sound classifier).
Then, we used the edit distance as an ordering function (i.e., the edit distance was calculated between each given name and the remaining candidates). 


\section{Results}
\label{sec:results}

\subsection{Performance Comparison}
\label{sec:performance_comparision}

In this section, we present the results of the experiments described in Section~\ref{sec:experimental_setup}.
The results of this evaluation are presented in Table~\ref{tab:top_10_performance}.

\textbf{SpokenName2Vec Evaluation.} 
In our evaluation, we assessed the performance of the \spokennametovec algorithm using several languages and accents; four configurations of \spokennametovec were developed using four languages (English, French, Spanish, and Italian) using Turi Create (TC), and one English was generated using pyAudioAnalysis (pyAA).
As seen in the first five rows of the Table~\ref{tab:top_10_performance}, most of the configurations performed similarly. 
For the accuracy measure, all of the configurations obtained average scores between 0.137 and 0.15.
The best configuration was the \spokennametovec algorithm that based its suggestions on Italian spoken names using Turi Create, which achieved an average accuracy score of 0.151.
The configurations that used Spanish, English, and French spoken names obtained similar high average accuracy scores of 0.148, 0.147, and 0.142, respectively.

For the F1 measure, the highest average scores were obtained by both configurations (Turi Create, and pyAudioAnalysis) when the English language was used (average F1 scores of 0.181 and 0.182, respectively). 
The configurations that used French, Spanish, and Italian obtained an average F1 of 0.175, 0.173, and 0.173, respectively.

For the precision measure, it can be seen that the highest average precision scores were obtained by the two English (pyAA and TC) and French \spokennametovec configurations which obtained an average precision@1 of 0.186, 0.184, and 0.183, respectively.
We can also see that as long as $k$ increases, the average precision@k decreases. 
The trend in similar performance among the leading configurations is also seen for the average precision@5 and average precision@10, although for precision@5 and average precision@10 the highest scores obtained by the Italian configuration with scores of 0.152 and 0.151, respectively for average precision@5 and average precision@10.
The configurations that used Spanish, French, and English using Turi Create had similar high performance.  

Regarding recall, in the table it can be seen that the highest average recall score was obtained by the configuration that used the English language and pyAudioAnalysis which had an average recall score of 0.169.
The next highest average recall scores were obtained using the French and English languages and Turi Create which had average recall scores of 0.133 and 0.13, respectively. 
The others had average recall scores of around 0.13.

\begin{table*}[t]
  \centering
  \caption{Performance obtained by various SpokenName2Vec configurations and other algorithms on Behind the Name Forenames dataset}
  \begin{tabular}{ccccccccc}
Method & Accuracy & F1 & AP@1 & AP@2 & AP@3 & AP@5 & AP@10 & Recall\\ [0.5ex] 
 \hline\hline
SpokenName2Vec TC (En) & 0.147 & 0.181 & 0.184 & \textbf{0.172} & \textbf{0.162} & 0.147 & 0.147 & 0.13  \\
SpokenName2Vec TC (Fr) & 0.142 & 0.175 & 0.183 & 0.167 & 0.158 & 0.149 & 0.143 & 0.133  \\
SpokenName2Vec TC (Sp) & 0.148 & 0.173 & 0.177 & 0.161 & 0.157 & 0.150 & 0.148 &  0.116 \\
SpokenName2Vec TC (It) & \textbf{0.151} & 0.173  & 0.165 & 0.16 & 0.157 & \textbf{0.152} & \textbf{0.151} & 0.113 \\
SpokenName2Vec pyAA (En) & 0.137 & \textbf{0.182} & \textbf{0.186} & 0.171 & 0.159 & 0.148 & 0.137 & 0.169 \\
\hline


GRAFT & 0.077 & 0.124 & 0.151 & 0.133 & 0.121 & 0.104 & 0.077 & \textbf{0.221} \\

\hline
Name2Vec & 0.021 & 0.037 & 0.079 & 0.063 & 0.052 & 0.038 & 0.021 & 0.075 \\ 
\hline
Soundex & 0.06 & 0.102 & 0.101 & 0.096 & 0.092 & 0.08 & 0.06 & 0.208  \\
Metaphone & 0.066 & 0.11 & 0.107 & 0.1 & 0.097 & 0.086 & 0.066 & 0.209 \\
DMetaphone & 0.068 & 0.112 & 0.107 & 0.102 & 0.098 & 0.088 & 0.068 & \textbf{0.221} \\
NYSIIS & 0.064 & 0.11 & 0.105 &	0.093 &	0.087 &	0.079 &	0.064 &	0.163 \\ 
MRA & 0.058 & 0.0919 &	0.093 &	0.086 &	0.082 &	0.073 &	0.058 & 0.144 \\ 
\hline
Jaro-Winkler & 0.044 & 0.077 &	0.076 &	0.075 &	0.071 &	0.061 &	0.044 & 0.177 \\ 
Edit Distance & 0.045 &	0.078 &	0.071 &	0.067 &	0.062 &	0.055 &	0.045 &	0.179\\ 
Damerau-Levenshtein & 0.046 & 0.08 & 0.071 & 0.065 & 0.062 & 0.056 &0.046 & 0.182 \\
\hline
  \end{tabular}
  \label{tab:top_10_performance}
\end{table*}

\subsubsection{Phonetic Encoding Algorithm Evaluation}

For the accuracy measure, we can see in Table~\ref{tab:top_10_performance} that all of the phonetic encoding algorithms obtain average scores around the value of 0.06; the highest score was obtained by Double Metaphone, with an average accuracy score of 0.068, and the lowest was obtained by MRA with an average score 0.058.
For the F1 measure, we can see that for all of these algorithms the average score was around 0.1.
Similarly, they all obtained an average precision@1 score of 0.1.
For recall, we can see that the phonetic encoding algorithms outperformed all of the other algorithms.
The highest average recall score was obtained by Double Metaphone, which had an average score of 0.221.
The second highest average recall scores were achieved by Metaphone and Soundex, with scores of 0.209 and 0.208, respectively.

\subsubsection{String Similarity Algorithm Evaluation}

As seen in Table~\ref{tab:top_10_performance}, the Jaro-Winkler, edit distance, and Damerau-Levenstein algorithms had similar performance on both accuracy measures, with average accuracy scores of 0.046, 0.45, and 0.44 and F1 scores of 0.08, 0.078, and 0.077, respectively.
For the precision measure, the Jaro-Winkler distance obtained the highest score for average precision@1 with a score of 0.076.
For the recall measure, the similarity algorithms obtained average scores second only to the phonetic encoding algorithms with average scores of 0.182, 0.179, and 0.177, respectively.

\subsubsection{Evaluation of GRAFT}

Table~\ref{tab:graft_comparision} presents the performance of \spokennametovec and GRAFT on forenames and surnames.
As can be seen, the proposed \spokennametovec algorithm achieved better performance on forenames and surnames than GRAFT in most of the aspects, except the recall measure.

Concerning forenames, we can see that GRAFT obtained an average accuracy score of 0.077 and an average F1 measure of 0.124 (see Table~\ref{tab:graft_comparision}).
For the precision metric, we can see that this algorithm obtained an average precision@1 score of 0.151 at its best. 
GRAFT obtained the highest recall score among all of the algorithms, with an average recall score of 0.221.

In contrast to GRAFT, we can see that \spokennametovec outperformed GRAFT: 
The average accuracy and F1 scores of the configuration of the English language and pyAudioAnalysis reached 0.137 and 0.182, respectively.
The highest average precision score reached at average precision@1 score of 0.186.
Regarding recall, GRAFT outperformed \spokennametovec with an average recall score of 0.221 as opposed to 0.169 reached by \emph{SpokenName2Vec}. 

The same trend can be seen when focusing on surnames.
GRAFT obtained an average accuracy score of 0.182 and average F1 score of 0.257 (see Table~\ref{tab:graft_comparision}).
The highest average precision score obtained was 0.514.
GRAFT's best average recall score was 0.867.

In contrast to \emph{GRAFT}, \spokennametovec obtained higher results, except recall measure: an average accuracy of 0.521, an average F1 of 0.563, an average precision@1 of 0.578, and an average recall of 0.667.

\subsubsection{Name2Vec Evaluation}

Table~\ref{tab:graft_comparision} also provides a comparison of the performance of \spokennametovec and Name2Vec.  
As can be seen, \spokennametovec outperformed Name2Vec on each measure. 
For the forename evaluation, Name2Vec obtained accuracy and F1 scores of 0.021 and 0.037, respectively, in contrast to \spokennametovec which obtained higher results (with accuracy and F1 scores of 0.137 and 0.182, respectively).
For precision, \spokennametovec obtained an average precision@1 score of 0.186, while Name2Vec obtained an average precision@1 score of 0.079.  
The same trend can be seen with respect to recall where \spokennametovec and Name2Vec obtained scores of 0.169 and 0.075, respectively.

Focusing on surnames, we can see that \spokennametovec outperformed Name2Vec again.
\spokennametovec obtained an average accuracy of 0.521, as opposed to 0.211 obtained by Name2Vec.
The highest precision of \spokennametovec was an average precision@1 score of 0.563, as opposed to the score of 0.211 achieved by Name2Vec.
On the recall measure, we can also see that \spokennametovec outperformed Name2Vec, with an average recall score of 0.667, as opposed to the score of 0.611 obtained by Name2Vec. 



\begin{table*}[ht]
  \centering
  \caption{Comparison of SpokenName2Vec, GRAFT, and Name2Vec}
  \begin{tabular}{ccccccccc}
Algorithm & Dataset & Type & Accuracy & F1 & AP@1 & AP@5 & AP@10 & Recall\\ [0.5ex] 
 \hline\hline
SpokenName2Vec & Behind the Name & Forenames & \textbf{0.137} & \textbf{0.182} & \textbf{0.186} & \textbf{0.148} & \textbf{0.137} & 0.169  \\
GRAFT & Behind the Name & Forenames & 0.077 & 0.124 & 0.151 & 0.104 & 0.077 & \textbf{0.221} \\
Name2Vec & Behind the Name &  Forenames & 0.021 & 0.037 & 0.079 & 0.038 & 0.021 & 0.075 \\
\hline
SpokenName2Vec & Ancestry & Surnames & \textbf{0.521} & \textbf{0.563} & \textbf{0.578} & \textbf{0.522} & \textbf{0.521} & 0.667  \\
GRAFT  & Ancestry & Surnames & 0.182  & 0.257 & 0.514 & 0.24 & 0.182 & \textbf{0.867} \\
Name2Vec & Ancestry & Surnames & 0.211 & 0.287 & 0.4 & 0.223 & 0.211 & 0.611 \\
  \end{tabular}
  \label{tab:graft_comparision}
\end{table*}


A similar picture can be seen for the surname evaluation, where \spokennametovec outperformed Name2Vec on every measure:
\spokennametovec obtained an average accuracy score of 0.521, while Name2Vec had an accuracy score of 0.211.
The same pattern is seen for F1, precision, and recall.

\subsection{Language and Accent Comparison}
\label{sec:language_and_accent_comparison}

Regarding the second research question which focuses on improving \emph{SpokenName2Vec}'s performance by determining the optimal language and associated accent, we found that the \spokennametovec configuration which used the spoken names in the French language was the most successful of the five configurations evaluated.
For 2,801 English names (commonly used in the United Kingdom and United States), the French \spokennametovec configuration obtained the highest precision score of 0.033 (see Table~\ref{tab:lang_models_comparision}). 
The configuration that came in second place was the English configuration with an average precision score of 0.025.
For 404 French names, the French configuration obtained the highest precision score of 0.03.
In second, third, and fourth place were the English, Spanish, and Russian configurations, which obtained average precision scores of 0.02, 0.018, and 0.015, respectively.
For 379 Spanish names and 307 Russian names, the French \spokennametovec configuration achieved first place with average precision scores of 0.03 and 0.06, respectively. 
Surprisingly, the \spokennametovec configuration that was the best for suggesting synonyms for 476 Italian names was the Spanish configuration, which obtained an average precision score of 0.028; the French configuration came next with an average precision score of 0.026.

\begin{table*}[t]
  \centering
  \caption{Performance of five configurations of the \emph{SpokenName2Vec} algorithm for suggesting synonyms}
  \begin{tabular}{ccccccc}
Usage & \shortstack{English \\ SpokenName2Vec} & \shortstack{French \\ SpokenName2Vec} & \shortstack{Spanish \\ SpokenName2Vec} & \shortstack{Italian \\ SpokenName2Vec} & \shortstack{Russian \\ SpokenName2Vec} \\ [0.5ex] 
 \hline\hline
English & 0.025 & \textbf{0.033} & 0.019 & 0.009 & 0.01   \\
French  & 0.02 & \textbf{0.03} & 0.018 & 0.009 & 0.015   \\
Spanish & 0.02 & \textbf{0.03} & 0.025 & 0.004 & 0.011  \\
Italian & 0.009 & 0.026 & \textbf{0.028} & 0.01 & 0.006 \\
Russian & 0.034 & \textbf{0.06} & 0.034 & 0.018 & 0.018 \\
  \end{tabular}
  \label{tab:lang_models_comparision}
\end{table*}

\begin{figure}[h!]
\centering
\includegraphics[scale=0.7]{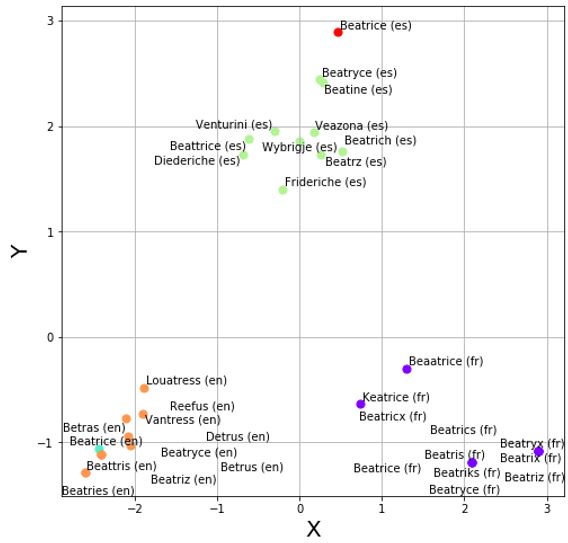}
\caption{Synonym distribution for the French name of Beatrice by the English, French, and Spanish \emph{SpokenName2Vec}'s configurations. The red and turquoise points represent the given name in the Spanish and English configurations, respectively. The green, purple and orange points represent the suggestions provided by the Spanish, French, and English configurations, respectively.}
\label{fig:beatrice}
\end{figure}

\begin{figure}[h!]
\centering
\includegraphics[scale=0.7]{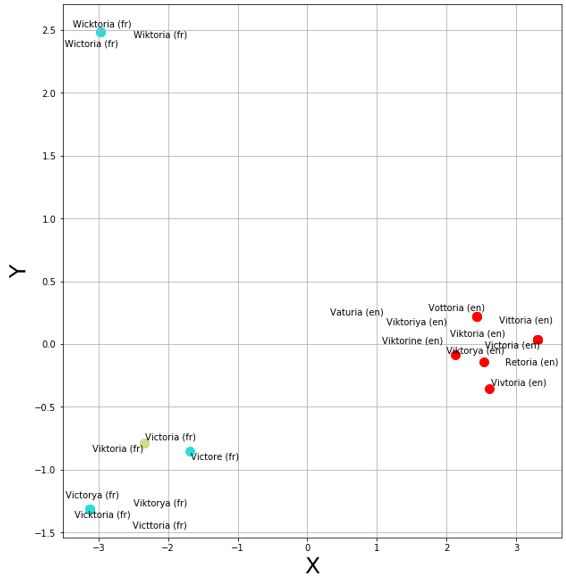}
\caption{Synonym distribution for the English name of Victoria by the English and French \emph{SpokenName2Vec} configurations. The green points represent the given name in the French configuration. The red and turquoise points represent the suggestions provided by the French,and English configurations, respectively.}
\label{fig:victoria}
\end{figure}

To illustrate the evaluation performed we chose two forenames: Beatrice, and Victoria. 
According to the Behind the Name website, the name of Beatrice is a commonly-used name in France for females and is probably derived from a feminine form of the Late Latin name Viator, which means voyager or traveler.\footnote{https://www.behindthename.com/name/be10atrice}
The name Victoria, meaning victory in Latin, was very rare in the English speaking world until the 19th century, when Queen Victoria began her long rule of the British Empire.\footnote{https://www.behindthename.com/name/victoria} 
We collected the candidates associated with each of the names, using all the five \spokennametovec configurations and a KNN classifier.
Then, for each name, we put the name and its associated candidates according to each language configuration on a vector space of two by applying dimensionality reduction using principal component analysis (PCA).
Doing so enables us to view the given name and its 10 associated candidates in multiple languages, as seen in Figures~\ref{fig:beatrice} and \ref{fig:victoria}.

Figure~\ref{fig:beatrice} presents the distribution of the given name and the names associated with it, provided by the \spokennametovec algorithm using three languages (English, French, and Spanish) for the given name of Beatrice.
In the figure, we can see that the French configuration was successful in suggesting four out of ten correct synonyms: Beatris (Russian), Beatrix (Dutch), Beatriz (Portuguese), and Beatryce (a rare form used by Americans and Brazilians).
The name of Beatriz (Portuguese) was detected as a true synonym by both the French and English \spokennametovec configurations, whereas the name of Beatryce (America and Brazil) was detected by all three configurations. 
It is interesting to note that the French configuration was also successful identifying six additional variations of the given name Beatrice that are not included on the website: Beaatrice, Beatricx, Beatrics, Beatryx, and  Beatriks.
Similarly, the English configuration identified the following names: Beatries and Beattris, and its Spanish counterpart found the names of Beattrice and Beatrich.

Figure~\ref{fig:victoria} presents the distribution of Victoria its associated names provided by the English and French \emph{SpokenName2Vec} configurations.
As can be seen, the French configuration successfully suggested two verified correct synonyms: Wiktoria (Polish) and Viktoria (German, Swedish, Norwegian, Danish, and many more). 
The English configuration was successful at suggesting the following synonyms: Vittoria (Italian) and Viktoriya (Bulgarian, Russian, and Ukrainian).
We can also see that the French configuration identified the following names which do not exist on the website: Wicktoria, Wictoria, Victorya, Viktorya, Vicktoria, and Victtoria.

\section{Discussion}
\label{sec:discussion}

Upon analyzing the results presented in Section~\ref{sec:results}, we can conclude the following:

First, the proposed novel \spokennametovec approach representing names based on automated speech had promising results and was found useful for the task of suggesting synonyms for a given name. 

Second, the suggested algorithm is generic.
For example, in the audio feature extraction step, features can be extracted using any available tool, and the algorithm does not depend on a single technique.
This was demonstrated by extracting audio features using two different tools: Turi Create and pyAudioAnalysis (see Section~\ref{sec:experimental_setup}); this finding suggests  that assessing other tools which capable converting the audio into a fixed-dimensional vector may help imporve \emph{SpokenName2Vec}'s results.
Our demonstration of this approach on forenames and surnames also demonstrated the approach's generality. 

Third, unlike many algorithms, such as Soundex and Name2Vec which support only the English language, the \spokennametovec algorithm supports multiple languages.
Its ability to extract valuable information based on speech without the necessity of working with text and grammar allows it to support many languages. 
This ability, which was demonstrated in our evaluation of the performance of different configurations that were used in English, Latin languages (such as French, Spanish, and Italian), and East Slavic languages (demonstrated using Russian) also shows the generality of the algorithm.

Fourth, with respect to performance on the task of suggesting synonyms, \spokennametovec was found superior to phonetic encoding and string similarity algorithms on all metrics (a difference found statistically significant using t-tests with $p-value < 0.05$) in terms of accuracy, F1, precision (see Table~\ref{tab:top_10_performance}).
For example, the \spokennametovec configuration of the English language and Turi Create obtained an average precision@1 score of 0.147, whereas Soundex and edit distance obtained scores of 0.06 and 0.045, respectively.  
Given this, we can conclude that the suggested name representation based on speech embedding is much more effective and accurate than the plain text code produced by phonetic encoding algorithms. 

Fifth, based on our comparison of \spokennametovec and Name2Vec, we conclude that \spokennametovec outperforms the Name2Vec approach presented by Foxcroft et al.~\cite{foxcroft2019name2vec}.
We base this conclusion on evaluation on forenames and surnames. 
The \spokennametovec algorithm was found to be superior on all metrics.
For instance, the average precision@1 of \spokennametovec was 0.186 as opposed to 0.079 obtained by Name2Vec.
A similar picture can be seen when evaluating surnames.
\spokennametovec obtained an average precision@1 score of 0.578 in contrast to Name2Vec's average score of 0.4.
The main disadvantage of the Name2Vec approach is related to its architecture. 
Name2Vec is a Doc2Vec model that relates to each name as a document and to each character that composes the given name as a word~\cite{foxcroft2019name2vec}.
This algorithm is limited  to suggest synonyms composed of only the characters of the given name.
Thus, it fails to suggest synonyms which include additional characters that do not exist in the given name.
For example, for the given name of Victoria, Name2Vec cannot suggest the associated correct synonym of Viktoria due to the absence of the character ``k'' in this given name.
Unlike Name2Vec, \spokennametovec does not depend on the characters, but rather depends on a similar sound.
Therefore, the absence of the character ``k'' is not an obstacle, and all of the \emph{SpokenName2Vec}'s configurations suggested the name Viktoria as a correct synonym for the given name of Victoria as demonstrated in Figure~\ref{fig:victoria}.

Sixth, it can also be seen that \spokennametovec is superior to GRAFT for forenames and surnames. (see Table~\ref{tab:graft_comparision}).
For example, the average F1 score obtained by \spokennametovec for suggesting synonyms for forenames was 0.182 as opposed to 0.124 obtained by GRAFT. 
It is important to understand that the algorithms are totally different from one another; 
\spokennametovec is capable of suggesting synonyms that sound alike but are written differently. 
GRAFT suggests synonyms based on historical ancestral relationships that are not necessarily related to sound.
We can therefore conclude from these results that for the task of synonym suggestion, it is more essential to utilize speech, machine and deep learning than historical data and network science. 
We believe that future research in which these two algorithms are combined should be very helpful and effective for this purpose.

Seventh, for the recall performance, we can see that GRAFT, and the phonetic encoding algorithms outperformed all other algorithms, including the proposed \spokennametovec algorithm. 
The highest average recall score (0.221) was obtained by both Double Metaphone and GRAFT. 
Next to them reached Metaphone, and Soundex with an average scores og 0.209, and 0.208, respectively.
The string similarity algorithms (edit and Damerau-Levenshtein distances) obtained average recall scores of 0.182, and 0.179, respectively.
The recall measure estimates the fraction of the total number of relevant names that were actually suggested.
Therefore, we deduce that these well-known algorithms can detect the largest number of correct synonyms in the long run, however, their mechanism misses many correct synonyms in the short-term (the top 10 suggestions). 
In contrast, \spokennametovec suggests synonyms with the highest likelihood first. 
This is the reason for its low recall scores.

Finally, with regard to the utilization of specific languages and accents for improving the \emph{SpokenName2Vec}'s performance, our initial assumption was that the best configuration of the \spokennametovec algorithm would be used the language of the targeted country or region, i.e., the English configuration would be the best for suggesting similar English names, the French configuration would be best for suggesting similar French names, etc. 
However, as can be seen, the \spokennametovec algorithm that used French speech was the best for suggesting synonyms for English, French, Spanish, and Russian forenames (see Table~\ref{tab:lang_models_comparision}). 
In addition, the French configuration was also very good at suggesting Italian names, however the best configuration for suggesting Italian names was the configuration that utilized the Spanish speech.
We can deduce that in the most cases, it is recommended to use the French configuration of \spokennametovec for improving synonym suggestion.

\section{Conclusion \& Future Work}
\label{sec:conclusion}

This paper introduces \emph{SpokenName2Vec}, a novel, generic multi-language algorithm which uses automated speech generation in different languages and accents and deep learning to address some of the challenges associated with synonyms.  
We provided a comprehensive description of our algorithm’s steps which start with the compilation of a collection of names using genealogical datasets; these datasets were used to generate audio segments reflecting the way humans pronounce the given names in several languages, such as English, French, Spanish, and Italian. 
Based on these speech segments, we extracted audio features, which serve as vector representations for each name.  
A supervised machine learning classifier was used to for finding the top 10 candidates most likely to be correct synonyms for a given name. 
Using a threshold, we filtered candidates that sound different from the given name and used an ordering function to retrieved the remaining names.
In this way, \spokennametovec was used to suggest synonyms for each given name in the ground truth.
We compared the performance of \spokennametovec on the task of suggesting similar name suggestion to the performance of 10 other search algorithms, including well-known phonetic encoding and string similarity algorithms, as well as GRAFT and Name2Vec, in our evaluation.
We make the following observations and conclusions:

The \spokennametovec algorithm was very useful for confronting the problem of suggesting synonyms for a given name, outperforming the other evaluated algorithms with respect to the accuracy, F1, and precision.   

The proposed algorithm is very generic. 
This is reflected in its demonstrated ability to (1) detect synonyms that sound alike but are written differently, a feature that shows its potential to support a  large number of languages, in contrast to other well-known algorithms (e.g., Soundex) that only support English;
(2) extract audio features using two different frameworks (Turi Create's sound classifier and pyAudioAnalysis), which shows \emph{SpokenName2Vec}'s ability to support various tools for feature extraction; and (3) use any supervised machine learning algorithm for name classification. 
The generality of this algorithm was also demonstrated in the suggestion of forenames and surnames.


Furthermore, our evaluation showed that the suggestions provided by all of the proposed \spokennametovec configurations are significantly better than the suggestions provided by all of the other algorithms evaluated, including GRAFT, Name2Vec, and phonetic encoding, and string similarity algorithms (a difference that was found to be statistically significant using t-tests with $p-value < 0.05$) for the performance measures of accuracy, F1, and precision.
Given this, we conclude that the proposed \spokennametovec algorithm should be used to solve the problem of suggesting synonyms.

With respect to recall, we can see that Double Metaphone and GRAFT obtained the highest average recall score of 0.221, as opposed to \spokennametovec which obtained an average score of 0.169.
We can deduce that GRAFT and Double Metaphone can detect the largest number of correct synonyms in the long run, however their mechanism misses many correct synonyms in the short-term (the top 10 suggestions). 
GRAFT's dependence on historical information for suggesting synonyms is an obstacle that the proposed \emph{SpokenName2Vec} algorithm doesn't face.


Our final conclusion is based on the results of language comparison (see Section~\ref{sec:language_and_accent_comparison}); we conclude that it is recommended to use the French configuration of \spokennametovec for improving synonym suggestion.
A possible future research direction is to examine other groups of names and datasets in order to understand the usefulness of the French configuration, as well as other configurations.
Another avenue to pursue is combining the sound and the family tree approaches to improve similar name suggestion. 

\section{Availability}

This study is reproducible research. 
Therefore, the Spoken Name dataset, as well as the algorithm for suggesting synonyms for a given name is available.\footnote{https://github.com/aviade5/SpokenName2Vec} 
Other datasets for evaluation are available upon request. 

\section{Acknowledgments}
The authors would like to thank the icons8 website (https://icons8.com) for their beautiful icons. 

\newpage\clearpage

\bibliographystyle{unsrt}
\bibliography{draft_for_arxiv.bib}
\newpage

\end{document}